\documentclass[10pt,twocolumn,letterpaper]{article}
\usepackage{titling}

\usepackage{cvpr}

\definecolor{cvprblue}{rgb}{0.21,0.49,0.74}
\usepackage[pagebackref,breaklinks,colorlinks,allcolors=cvprblue]{hyperref}
\usepackage{makecell}

\def\paperID{9} % *** Enter the Paper ID here
\def\confName{CVPR}
\def\confYear{2025}

\makeatletter
\def\section{\@startsection {section}{1}{\z@}%
  {0.5\baselineskip plus 0.1\baselineskip minus 0.1\baselineskip}%
  {0.3\baselineskip plus 0.1\baselineskip minus 0.1\baselineskip}%
  {\normalfont\large\bfseries}}
\makeatother

\begin{document}
\title{Harnessing the Power of Training-Free Techniques in Text-to-2D Generation \\ for Text-to-3D Generation via Score Distillation Sampling
}

\author{
Junhong Lee \quad Seungwook Kim \quad Minsu Cho \vspace{1.5mm} \\
Pohang University of Science and Technology (POSTECH), South Korea \vspace{1.5mm} \\
}

\maketitle
\pagestyle{plain}
\begin{abstract}
Recent studies show that simple training-free techniques can dramatically improve the quality of text-to-2D generation outputs, \textit{e.g.,} Classifier-Free Guidance (CFG) or FreeU.
However, these training-free techniques have been underexplored in the lens of Score Distillation Sampling (SDS), which is a popular and effective technique to leverage the power of pretrained text-to-2D diffusion models for various tasks.
In this paper, we aim to shed light on the effect such training-free techniques have on SDS, via a particular application of text-to-3D generation via 2D lifting. 
We present our findings, which show that varying the scales of CFG presents a trade-off between object size and surface smoothness, while varying the scales of FreeU presents a trade-off between texture details and geometric errors.
Based on these findings, we provide insights into how we can effectively harness training-free techniques for SDS, via a strategic scaling of such techniques in a dynamic manner with respect to the timestep or optimization iteration step.
We show that using our proposed scheme strikes a favorable balance between texture details and surface smoothness in text-to-3D generations, while preserving the size of the output and mitigating the occurrence of geometric defects.
\end{abstract}    
\section{Introduction}
\label{sec:intro}

Diffusion models are now the de-facto standard for image generation, due to their ability to generate a wide variety of high-quality, realistic images~\cite{dhariwal2021diffusionmodelsbeatgans, esser2021imagebartbidirectionalcontextmultinomial, gal2022imageworthwordpersonalizing, gu2022vectorquantizeddiffusionmodel, ho2020denoisingdiffusionprobabilisticmodels, 
% kawar2023imagictextbasedrealimage, 
kumari2023multiconceptcustomizationtexttoimagediffusion, nichol2022glidephotorealisticimagegeneration}. 
% Unlike Autoencoders~\cite{bank2021autoencoders}, which have fast training speeds but produce lower-quality images, or GANs~\cite{goodfellow2014generativeadversarialnetworks}, which can generate high-quality images but struggle to produce diverse results, diffusion excels in both quality and diversity. 
A popular application of diffusion models is conditional generation, where the most prominent case is to generate images conditioned on the input text prompts \textit{i.e.,} text-to-2D generation~\cite{nichol2022glidephotorealisticimagegeneration, ramesh2021zeroshottexttoimagegeneration, ramesh2022hierarchicaltextconditionalimagegeneration, saharia2022paletteimagetoimagediffusionmodels, saharia2022photorealistictexttoimagediffusionmodels, yu2022scalingautoregressivemodelscontentrich,saharia2021imagesuperresolutioniterativerefinement, rombach2022highresolutionimagesynthesislatent}.
To yield outputs with improved quality and controllability, large number of studies endeavor to expand the diversity of the training data~\cite{georgiev2023journeydestinationdataguides, zheng2024intriguingpropertiesdataattribution, schuhmann2022laion5bopenlargescaledataset, chen2020wavegradestimatinggradientswaveform, ho2022videodiffusionmodels, kong2021diffwaveversatilediffusionmodel}, propose novel training techniques~\cite{karras2024analyzingimprovingtrainingdynamics, kim2022softtruncationuniversaltraining}, or propose improved architectures for the diffusion models~\cite{pernias2023wuerstchenefficientarchitecturelargescale, zhang2024improvingefficiencydiffusionmodels, deja2022analyzinggenerativedenoisingcapabilities, balaji2023ediffitexttoimagediffusionmodels, lee2023multiarchitecturemultiexpertdiffusionmodels, go2023addressingnegativetransferdiffusion}.

Apart from the above approaches, a body of recent work propose training-free techniques to enhance the output quality of text-to-2D.
For example, Classifier-Free Guidance (CFG)~\cite{ho2022classifierfreediffusionguidance} aims to improve sample quality by combining conditional and unconditional sampling without the need for an explicit classifier, allowing the model to better reflect specific conditions during generation.
More recently, FreeU~\cite{si2023freeufreelunchdiffusion} scales each of the backbone features and skip features in the upsampling layers of the UNet, allowing the model to capture details and improve the image quality without additional computational costs.

Pretrained text-to-2D diffusion models serve as the foundation for various other computer vision tasks as well, including dense prediction tasks such as image matching~\cite{zhang2023talefeaturesstablediffusion, luo2024diffusionhyperfeaturessearchingtime, tang2023emergentcorrespondenceimagediffusion, hedlin2023unsupervisedsemanticcorrespondenceusing} or
image segmentation~\cite{baranchuk2022labelefficientsemanticsegmentationdiffusion, amit2022segdiffimagesegmentationdiffusion, tan2023diffssdiffusionmodelfewshot}, and also other high-level tasks such as image
super-resolution~\cite{saharia2021imagesuperresolutioniterativerefinement, li2021srdiffsingleimagesuperresolution}, semantic editing~\cite{liu2024dragnoiseinteractivepointbased, voynov2023pextendedtextualconditioning}, image inpainting~\cite{lugmayr2022repaintinpaintingusingdenoising, grechka2023gradpaintgradientguidedinpaintingdiffusion}, or colorization~\cite{song2021scorebasedgenerativemodelingstochastic, zabari2023diffusingcolorsimagecolorization, liu2023improveddiffusionbasedimagecolorization}.
A particularly interesting application of text-to-2D diffusion models is text-to-\textbf{\textit{3D}} generation via 2D lifting, which was first proposed by DreamFusion~\cite{poole2022dreamfusiontextto3dusing2d}.
The process of 2D lifting is facilitated by Score Distillation Sampling (SDS), which leverages the text-to-2D diffusion model as a \textit{critique} to optimize a 3D representation such as NeRF~\cite{mildenhall2020nerfrepresentingscenesneural} or NeUS~\cite{wang2023neuslearningneuralimplicit} from scratch.

In this paper, we identify that the effects of training-free techniques - which show dramatic effects on text-to-2D generation - have been underexplored in the lens of SDS.
To this end, we explore how such training-free techniques affect the text-to-3D generation process.
Our findings show that while the application of such training-free techniques to score distillation is effective, \textit{how} we apply these techniques holds higher importance.
Unlike application to text-to-2D diffusion that focused narrowly on scaling for high sample quality, application to score distillation has certain trade-offs based on the magnitude of the scaling factor with respect to the timestep of diffusion or optimization iteration of 3D representation.
To effectively harness the power of such training-free techniques while overcoming the trade-offs associated with its application to SDS, we devise strategic dynamic scaling of such techniques based on the insight that FreeU adjusts features within the diffusion process while CFG regulates the diffusion scores.

In summary, the contributions of our work are threefold:
\begin{enumerate}
    \item We analyse the effect of training-free techniques \textit{e.g.,} FreeU and CFG, on score distillation, in the particular context of text-to-3D generation via 2D lifting.
    \item Based on the above observations, we devise a well-motivated scaling scheme in a dynamic manner, to maximize the potential of FreeU and CFG in enhancing the quality of text-to-3D generation outputs.
    \item We demonstrate the efficacy of our dynamic scaling scheme across various SDS-based pipelines, validated by strong qualitative comparisons and a user study.
\end{enumerate}

\section{Related work}
\label{sec:relatedwork}
\smallbreak
\noindent
\textbf{Text-to-2D Generation via diffusion.}
Diffusion models~\cite{ho2020denoisingdiffusionprobabilisticmodels} have recently demonstrated superior performances in text-to-2D generation.
Stable Diffusion~\cite{rombach2022highresolutionimagesynthesislatent} first facilitated text-conditioned image generation by sending images to a latent space, which also enabled the efficient manipulation of high-resolution images by enabling the processing of a larger amount of information within the same memory constraints.
This work motivated many strong image-to-2D models including 
DALL-E2~\cite{ramesh2022hierarchicaltextconditionalimagegeneration} and Imagen~\cite{saharia2022photorealistictexttoimagediffusionmodels}.
Text-to-2D models aim to generate the desired image from text, and are used for various tasks such as image analysis~\cite{zhang2023talefeaturesstablediffusion, luo2024diffusionhyperfeaturessearchingtime, tang2023emergentcorrespondenceimagediffusion, hedlin2023unsupervisedsemanticcorrespondenceusing, baranchuk2022labelefficientsemanticsegmentationdiffusion, amit2022segdiffimagesegmentationdiffusion, tan2023diffssdiffusionmodelfewshot}, and image modification~\cite{meng2022sdeditguidedimagesynthesis, liu2024dragnoiseinteractivepointbased, voynov2023pextendedtextualconditioning, lugmayr2022repaintinpaintingusingdenoising, grechka2023gradpaintgradientguidedinpaintingdiffusion}. 

In this work, we focus on improving the output quality of text-to-3D models that use 2D lifting, which builds on the strong capabilities of text-to-2D diffusion models via the means of Score Distillation Sampling~\cite{poole2022dreamfusiontextto3dusing2d}.

\smallbreak
\noindent
\textbf{Text-to-3D Generation.}
3D generative models can be trained on explicit representations~\cite{wu2017learningprobabilisticlatentspace, chen2018text2shapegeneratingshapesnatural, yang2019pointflow3dpointcloud, cai2020learninggradientfieldsshape, zhou20213dshapegenerationcompletion}, or can also be based on implicit representations~\cite{mildenhall2020nerfrepresentingscenesneural, wang2023neuslearningneuralimplicit}. 
Recent advances in text-to-3D generation have adopted a groundbreaking approach known as Score Distillation Sampling (SDS) to facilitate text-to-3D generation via 2D lifting~\cite{poole2022dreamfusiontextto3dusing2d, lin2023magic3dhighresolutiontextto3dcontent, shi2024mvdreammultiviewdiffusion3d, yu2023textto3dclassifierscoredistillation, seo2024let2ddiffusionmodel, hong2023debiasingscoresprompts2d, chen2023fantasia3ddisentanglinggeometryappearance, wang2023prolificdreamerhighfidelitydiversetextto3d, wang2022scorejacobianchaininglifting, metzer2022latentnerfshapeguidedgeneration3d, cohenbar2023setthescenegloballocaltraininggenerating, kim2024enhancing3dfidelitytextto3d, haque2023instructnerf2nerfediting3dscenes, tsalicoglou2023textmeshgenerationrealistic3d}.
DreamFusion~\cite{poole2022dreamfusiontextto3dusing2d} first proposed SDS for text-to-3D synthesis by optimizing NeRF~\cite{mildenhall2020nerfrepresentingscenesneural} MLP parameters using a pre-trained text-to-2D diffusion model as the critique. 
Magic3D~\cite{lin2023magic3dhighresolutiontextto3dcontent} 
takes a coarse-to-fine strategy that utilizes both low-resolution and high-resolution diffusion priors for improved efficiency.
However, using a single-view 2D text-to-image model as the diffusion prior poses 3D consistency problems, leading to the Multi-face Janus or content drift problems.
To alleviate the multi-view consistency problem, MVDream~\cite{shi2024mvdreammultiviewdiffusion3d} uses a multi-view diffusion as the prior, which handles images from various angles simultaneously to tackle the issue of 3D consistency. 

In this work, we aim to delve into the underexplored direction of applying training-free techniques to improve the output quality of SDS-based text-to-3D generation methods \textit{without} additional training.

\smallbreak
\noindent
\textbf{Techniques to improve the results of Text-to-\{2D,3D\} Generation.}
A prominent way to improve the generation quality is by providing guidance.
Classifier Guidance (CG)~\cite{dhariwal2021diffusionmodelsbeatgans} provides guidance by mixing the scores from training an additional classifier and the origin. 
Classifier-Free Guidance(CFG)~\cite{ho2022classifierfreediffusionguidance}, on the other hand, proposes that similar results can be achieved without a classifier by mixing the score estimates of an unconditional diffusion model trained with the conditional diffusion model.
To enable guidance in a non-conditional setting, Self-Attention Guidance(SAG)~\cite{hong2023improvingsamplequalitydiffusion} proposes to apply Gaussian Blur to patches with high attention scores at each timestep, interpreting the patches as containing high-frequency information.
Perturbed-Attention Guidance (PAG)~\cite{ahn2024selfrectifyingdiffusionsamplingperturbedattention} adopts a method of perturbing only the self-attention map to prevent excessive deviation from the original sample, helping the model to focus on prominent points at each timestep.

Another way to improve the generation quality is by \textit{scaling} the information from different modules or elements of the network pipeline.
Consistent123~\cite{lin2024consistent123imagehighlyconsistent} demonstrates that a large CFG scale contributes to the geometry of synthesized object views while a low CFG scale helps to refine the texture details. 
FreeU~\cite{si2023freeufreelunchdiffusion} proposes to enhance the backbone features in the upsampling layers of the diffusion UNet to improve image quality and attenuate the low-frequency part of the skip features, in order to prevent oversmoothing caused by amplification.

In this work, we determine the effect of training-free techniques on text-to-3D generation via 2D lifting, and propose a novel scheme to effectively harness the power of such techniques to improve the 3D generation quality.
\section{Preliminary}
\label{sec:preliminary}
\subsection{FreeU}
FreeU analyzes and scales the denoising process of Denoising Diffusion Probabilistic Models (DDPM) to improve the quality of generation.
In the upsampling layers of the U-Net, the backbone feature and the skip feature are concatenated; various studies indicate that in this process, the concatenated backbone feature contributes directly to the denoising process, while the skip feature provides high-frequency information to the decoder module.

In FreeU, the backbone features are amplified using a structure-related scaling method, which dynamically adjusts the scaling of backbone features for each sample. 
The backbone feature scaling is applied to only half of  the channels to mitigate oversmoothing, as follows:
\begin{equation}\label{eq:FreeUbackbone}
    x'_{l,i} = 
        \begin{cases}
            x_{l,i} \odot b_l, & \text{if }i < C/2\\
            x_{l,i},                & \text{otherwise}
        \end{cases}
\end{equation}
where $x_{l,i}$ represents the $i$-th channel of the feature map, $b_l$ is a scaling factor for backbone feature. 
Indeed, the FreeU proposes a method of scaling by simply adjusting the backbone factor from 1 to $b_l$. 

The skip feature diminishes the low-frequency component of the features, aiming to mitigate the over-smoothing of textures caused by the enhanced denoising capability from the scaled backbone features. 
This approach follows a method of scaling the component smaller than a threshold frequency using the Fourier transform.
\begin{align*} 
    \mathcal{F}(h_{l,i}) &= FFT(h_{l,i})\\
    \mathcal{F}'(h_{l,i}) &= \mathcal{F}(h_{l,i}) \odot \beta_{l,i}\\
    h'_{l,i} &= IFFT(\mathcal{F}'(h_{l,i}))
\end{align*} 
\begin{equation}\label{eq:FreeUskip}
    \text{,where }\beta_{l,i}(r) = 
                    \begin{cases}
                        s_l, & \text{if }r<r_{threshold}\\
                        1 & \text{otherwise}
                    \end{cases}
\end{equation}

\subsection{Classifier-Free Guidance}
Classifier Guidance modifies the diffusion score $\epsilon_\theta (z_\lambda , c) \approx -\sigma_\lambda \nabla_{z_\lambda} \log p(z_\lambda | c)$ function to include the gradient of the log-likelihood of an auxiliary classifier model, helping the generative model produce images corresponding to the correct label, as follows:
\begin{equation}
    \Tilde{\epsilon_\theta}(z_\lambda , c) = \epsilon_\theta (z_\lambda , c) - \omega \sigma_\lambda \nabla_{z_\lambda} \log p_\theta (c|z_\lambda)
\end{equation}
The classifier part of CG has the following relationship by Bayes' rule: $p(c|z_\lambda) \propto p(z_\lambda | c) / p(z_\lambda)$. If we had access to exact scores $\epsilon^*(z_\lambda , c)$ and $\epsilon^*(z_\lambda)$ (of $p(z_\lambda |c)$ and $p(z_\lambda)$, respectively), then the gradient of the implicit classifier would be $\nabla_{z_\lambda} \log p (c|z_\lambda) = -\frac{1}{\sigma_\lambda} \left[\epsilon^*(z_\lambda , c) - \epsilon^*(z_\lambda) \right]$. Ultimately, Classifier-Free Guidance demonstrate that simply conditioning and unconditioning on the same structure can achieve effects similar to classifier guidance, as follows:
\begin{equation}
    \Tilde{\epsilon_\theta}(z_\lambda , c) = (1+\omega) \epsilon_\theta (z_\lambda , c) - \omega \epsilon_\theta (z_\lambda)
\end{equation}

\subsection{Score Distillation Sampling}
Score distillation is an idea first proposed in DreamFusion, which uses the \textit{score} from diffusion to optimize NeRF:
\begin{equation}\label{eq:diffusion_loss}
    \mathcal{L}_{\text{Diff}}(\phi, \mathbf{x}) = \mathbb{E}_{t, \epsilon} 
    \left[ w(t) \ \|\epsilon_{\phi}(\alpha_t \mathbf{x} + \sigma_t \epsilon; t) - \epsilon \|_2^2 \right]
\end{equation}
The loss function of a diffusion model can be expressed as a simplified form of the evidence lower bound (ELBO) during the training process, which leads to a weighted denoising score matching objective, optimizing the parameters $\phi$.
\cref{eq:diffusion_loss} represents the loss of reparameterized diffusion model, where $w(t)$ is a weighting function that depends on the timestep $t$, and ${t \sim \mathcal{U}(0,1)}$, ${\epsilon \sim \mathcal{N}(0,\mathbf{I})}$.
\begin{equation}\label{eq:sds_loss}
    \nabla_{\theta} \mathcal{L}_{\text{SDS}}
    \triangleq \mathbb{E}_{t, \epsilon} 
    \left[ w(t) \left( \hat{\epsilon}_{\phi}(z_t; y, t) - \epsilon \right) \frac{\partial g(\theta)}{\partial \theta} \right]
\end{equation}
\cref{eq:sds_loss} represents the gradient used to update the NeRF paremeter $\theta$ in score distillation sampling.
In this method, instead of an input image being used in the original diffusion loss \cref{eq:diffusion_loss}, an image (NeRF rendering) generated by a differentiable generator $g$ is used. 

Ultimately, the SDS loss guides the value generated by NeRF to align with the value produced by the diffusion prior conditioned on the text embedding. 
The types of embeddings, the diffusion model used as the prior, and the type of 3D representation can be interchangeable.
\section{Method}
\label{sec:method}

\noindent
\textbf{Overview.}
We first investigate the effects of FreeU on text-to-multi-view diffusion, as multi-view diffusion priors have been shown to outperform single-view diffusion priors. 
Building on this analysis, we interpret the effects of multi-view diffusion on score distillation to devise a dynamic scaling strategy of FreeU to maximize its positive effects.
Similarly, we explore the differences between CFG in the context of text-to-2D generation and score distillation, and present a scaling strategy to maximize the positive impact drawn from CFG. 
Finally, we offer an integrated, cohesive approach to harmoniously harness each training-free technique on SDS.

\subsection{FreeU dynamic scaling on score distillation.}

\smallbreak
\noindent
\textbf{FreeU on text-to-2D diffusion.}
FreeU's~\cite{si2023freeufreelunchdiffusion} experiments indicate that as the backbone features are strengthened, high-frequency components are suppressed.
However, since these experiments were conducted within a text-to-single-view diffusion framework, it is not inherently expected that the same effect will manifest in score distillation.
To this end, we explore the effect of FreeU in score distillation based solely on the fact that the backbone scaling factor, a fundamental interpretation of diffusion, enhances the denoising capability of the UNet.

\begin{figure}
    \centering
    \includegraphics[width=\linewidth]{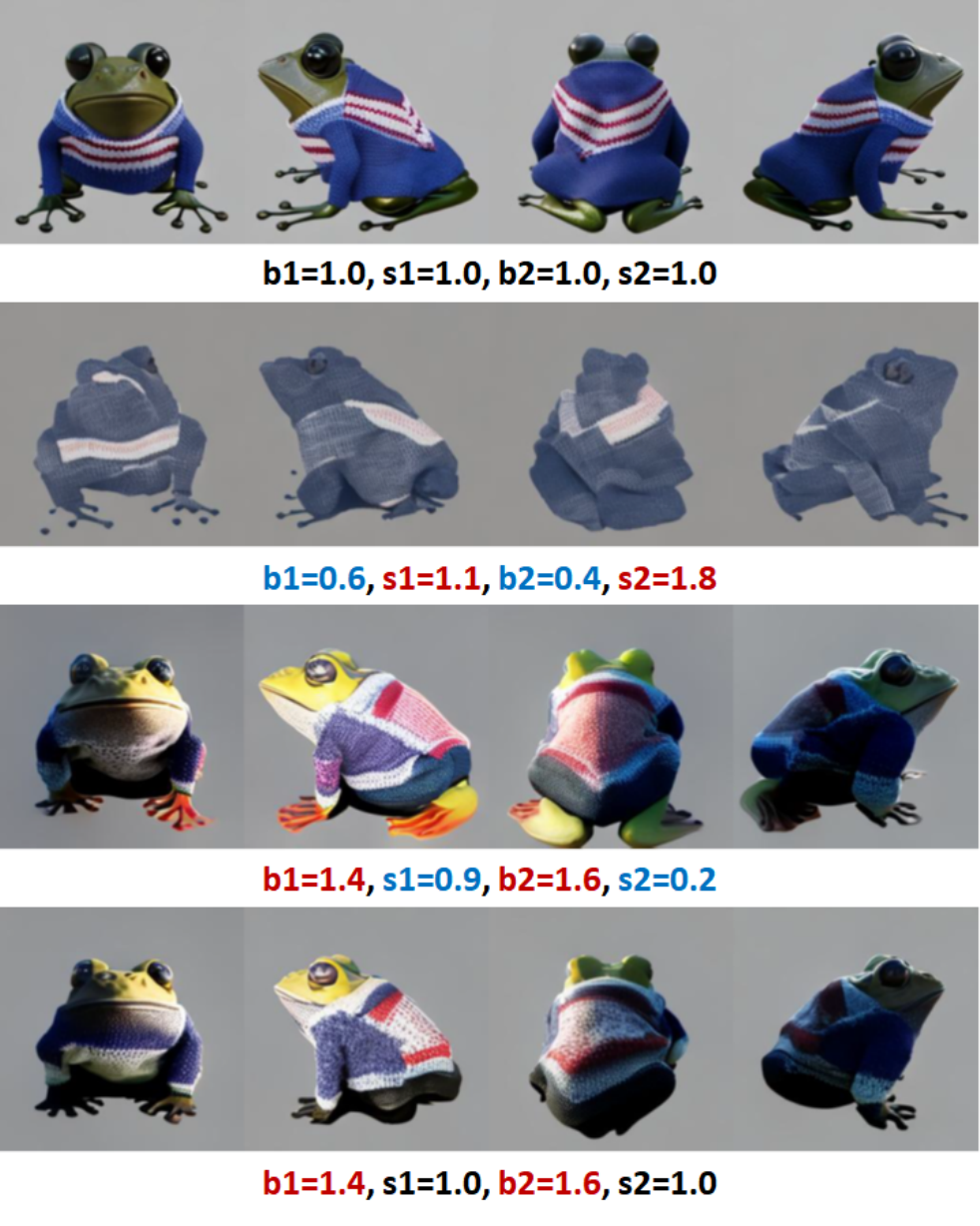}
    \caption{
    \textbf{FreeU on multi-view diffusion.} 
    Results for the prompt '\textit{a DSLR photo of a frog wearing a sweater, 3D asset.}' 
    From bottom to top: scaling only backbone features, scaling both backbone and skip features, scaling with values symmetric to $1.0$, and no scaling. 
    Each scaling follows values suggested by FreeU.
    }
    \label{fig:FreeU_multiviewdiffusion}
\end{figure}

\smallbreak
\noindent
\textbf{FreeU on multi-view diffusion.}
Before examining the effects of training-free techniques on SDS, it is important to understand how FreeU influences the multi-view diffusion used as prior. 
As illustrated in \cref{fig:FreeU_multiviewdiffusion}, skip feature scaling has minimal impact on the quality of the results, and amplifying the backbone features enhances the denoising capability of UNet, resulting in outputs with well-depicted details. 
However, the enhancement of backbone features also increases the risk of inconsistency in multi-view diffusion\footnote{We guide the readers to the supplementary for visual examples}, posing a critical trade-off. 

\begin{figure*}[ht]
    \centering
    \includegraphics[width=\linewidth]{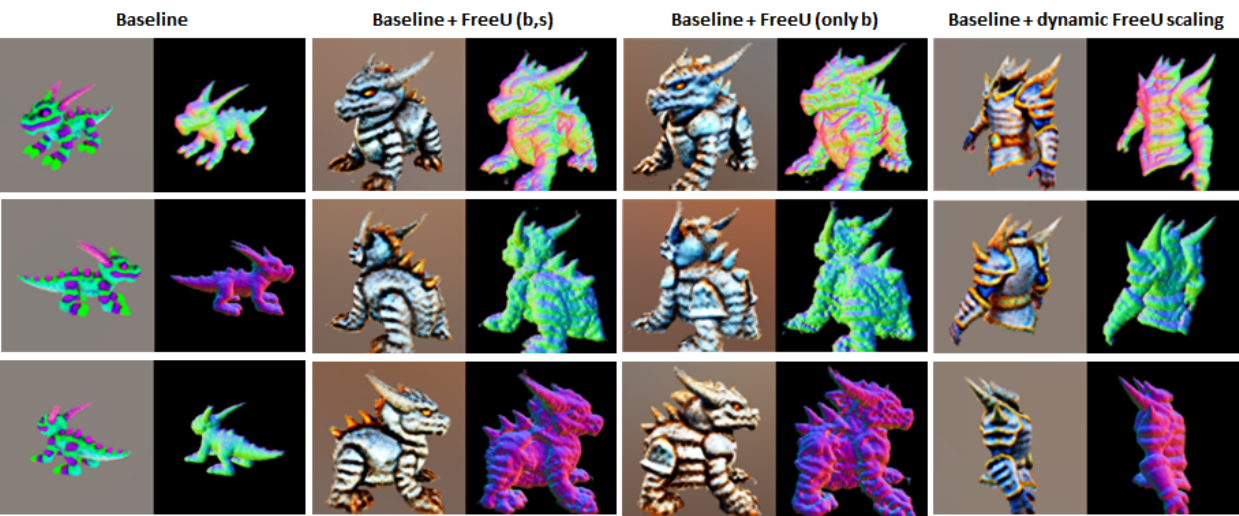}
    \caption{
    \textbf{FreeU on Score Distillation Sampling}
    These are generated from the prompt 'Dragon armor, 3D asset.' 
    While FreeU scaling enhances detail capture in SDS, it also increases the risk of geometric defects, such as body distortion as above. 
    Our proposed dynamic scaling technique preserves detail while avoiding geometric issues. 
    Similar to its effect in diffusion, FreeU scaling in SDS shows that skip feature scaling does not significantly contribute to quality improvement.
    }
    \label{fig:FreeU_SDS}
\end{figure*}

\smallbreak
\noindent
\textbf{FreeU on score distillation.}
Based on the interpretations presented above, we analyze the relationship between the effects of FreeU in multi-view diffusion and its impact on score distillation via multi-view diffusion. 
As shown in \cref{fig:FreeU_SDS}, the amplification of backbone features by FreeU significantly improves the detail representations but also increases the likelihood of geometric defects or artifacts, and the skip feature scaling does not have much impact on the results as in multi-view diffusion.
Similar to its application in multi-view diffusion, the FreeU technique involves a trade-off between improved detail and an increased risk of geometric errors.
From these consistent results, we speculate that the risk of inconsistency in multi-view diffusion is associated with the risk of geometric error occurrence in the iterative optimization of score distillation.

\smallbreak
\noindent
\textbf{Dynamic scaling technique of FreeU}
In 2D lifting, score distillation via multi-view diffusion involves annealing the timesteps of diffusion, which helps to make the shape more complete~\cite{shi2024mvdreammultiviewdiffusion3d}. 
During optimization, the early stage with large timestep is used to focus on generating high-level geometry, and the late stage with small timestep is employed to render texture details. 
The enhancement of texture details, a key advantage of FreeU, is driven by the amplification of backbone features during the later stages via small timestep. 
From the observations, we propose a dynamic scaling method that adjusts the FreeU scaling factor according to the \textit{timestep}, aiming to capture the detail while suppressing geometric flaws or artifacts.
We recommend a dynamic scaling with $b_t < 1$ when timestep $t$ is large to stabilize the object's shape, and $b_t > 1$ when $t$ is small to increase sensitivity to detail (\cref{fig:FreeU_SDS}), where $b_t$ is the backbone feature scaling factor for each timestep $t$.
The proposed dynamic scaling method achieves an appropriate balance between the absence of scaling and the application of FreeU without any dynamic scaling.

\subsection{CFG dynamic scaling on score distillation}

\smallbreak
\noindent
\textbf{CFG on text-to-2D.}
Since CFG does not modify features within the diffusion UNets, we do not need to assess its impact on multi-view diffusion. 
However, in score distillation, where the diffusion score is used to optimize the 3D representation, it is crucial to analyze whether CFG has the same effect as it does in text-to-2D diffusion \textit{i.e.,} ensuring that the output accurately reflects the text.

\begin{figure*}
    \centering
    \includegraphics[width=\linewidth]{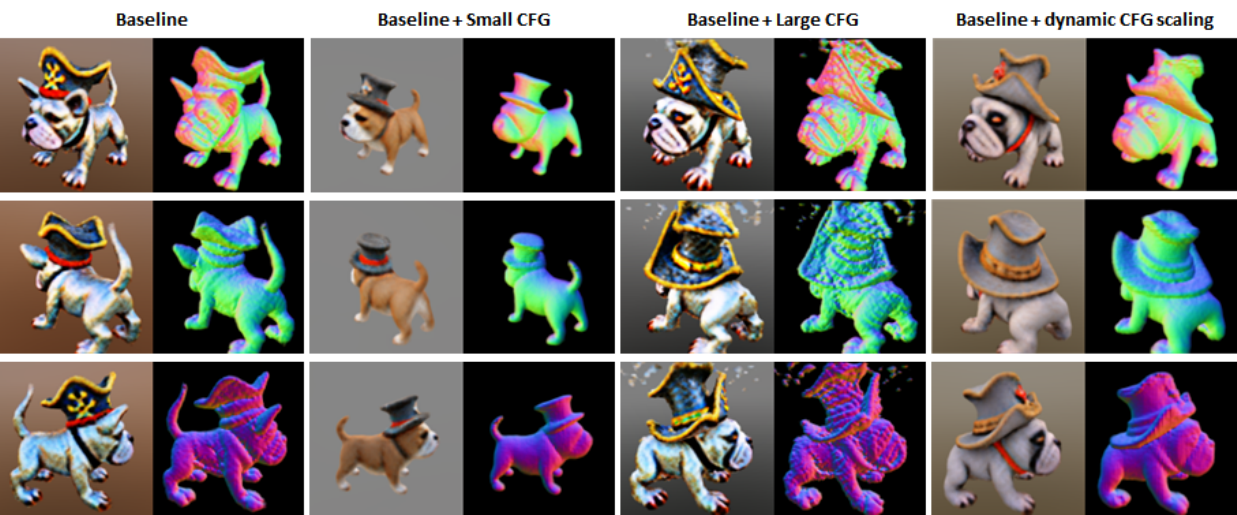}
    \caption{
    \textbf{CFG on Score Distillation Sampling}
    The guidance weights, which serve as CFG scaling variables, are 50(origin), 10(small CFG), 100(large CFG), and 100 to 10(dynamic scaling) from left to right. 
    Scaling with larger values increases the object's size and roughens the surface, and vice versa. 
    Our dynamic scaling technique allows the object to maintain its size while achieving a surface smoothness comparable to when the guidance weight is 10.
    }
    \label{fig:CFG_SDS}
    \vspace{-3.0mm}
\end{figure*}

\smallbreak
\noindent
\textbf{CFG on score distillation.}
As illustrated in \cref{fig:CFG_SDS}, CFG on SDS has a trade-off between surface smoothness and size of the object, where larger guidance weight leads to larger sizes and rougher surfaces.
Increasing the guidance weight causes the rendered view from the 3D representation to diverge farther away from the diffusion prior, leading to a higher loss. 
During optimization, this divergence causes the object to be larger in the early stages, where high-level geometry is formed, and a rougher surface in the later stages, where texture details are refined. 
The increase in object size indicates an improvement in the model's generative capability, but the rough surface involves the presence of geometric defects. 
Therefore, scaling CFG dynamically is appropriate to gain benefits both in terms of size and smoothness.

\smallbreak
\noindent
\textbf{Dynamic scaling technique of CFG.}
Since CFG operates outside the diffusion process by scaling the score, it is more appropriate to schedule it according to the step of the iterative optimization rather than the timestep $t$.
While there is a tendency where the timestep decreases as the iteration step increases, timestep annealing involves random selection within a particular range for each stage, making a clear distinction between the following iteration and timestep. 
To fully leverage the advantages of CFG in SDS, 
we propose to dynamically scale the guidance weight according to the iteration as shown in \cref{fig:CFG_SDS}. 
The dynamic scaling takes the following rules: during the later stages, which handle high-frequency details, we reduce the guidance weight for surface smoothness. 
During the earlier stages of optimization, where low-frequency geometries are determined, we increase the guidance weight to avoid downsizing the object size from the lower guidance values in the later stages.

\subsection{Simultaneous application of FreeU and CFG dynamic scaling}\label{sec:simultaneous}

\begin{figure*}
    \centering
    \includegraphics[width=0.825\textwidth]{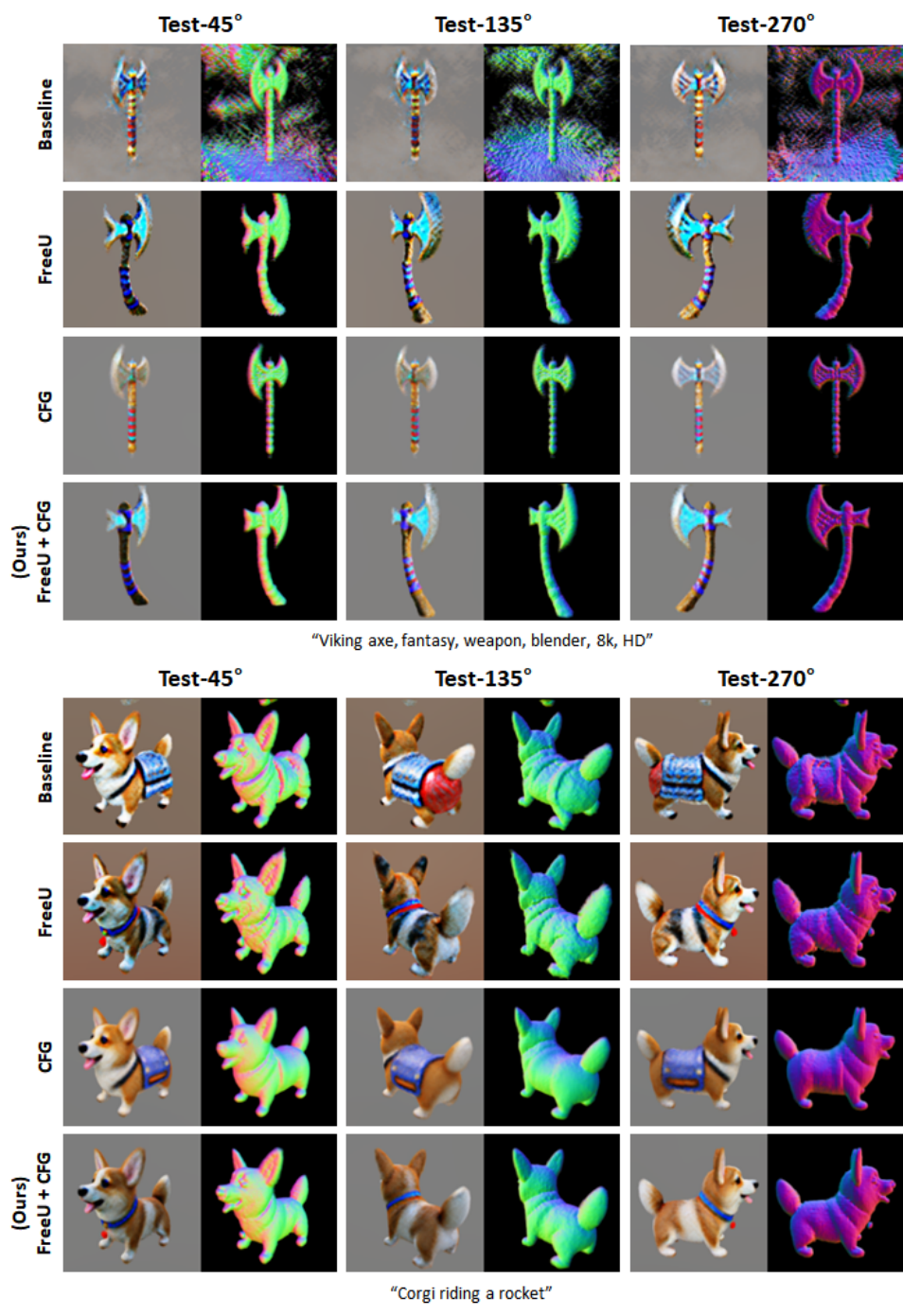}
    \caption{
    \textbf{Samples generated by score distillation with or without dynamic scaling training-free techniques.}
    }
    \label{fig:experiment}
\end{figure*}

In score distillation, FreeU dynamic scaling is specifically designed to capture the details of the output and suppress geometric flaws and artifacts, whereas CFG dynamic scaling focuses on increasing the object size and smoothing its surface, as depicted in \cref{fig:experiment}. 
Although there is no guarantee that these two dynamic scaling methods influence the results independently, they are applied at different points in the score distillation process and serve distinct primary roles.
Additionally, investigations through experiments with various dynamic scaling combinations suggest that their interdependence is likely minimal.
\section{Experiments}
\label{sec:experiments}
\subsection{Implementation details}
We primarily conduct experiments using MVDream ~\cite{shi2024mvdreammultiviewdiffusion3d}, a representative open-source implementation of multi-view diffusion, as our baseline text-to-3D model.
Consistently with MVDream, we use NeRF~\cite{mildenhall2020nerfrepresentingscenesneural}, as the 3D representation; we optimize it over 10,000 iteration steps with timestep annealing.
Due to GPU memory constraints, we use NeRF rendering sizes of 32x32 for the first 5,000 steps, and then increase it to 100x100 for the next 5,000 steps, unlike the original settings of 64x64 to 256x256. 

Our dynamic scaling rules are accompanied by the following values:
The parameters for FreeU ($b1$,$s1$,$b2$,$s2$) are linearly mapped from ($0.6, 1.1, 0.4, 1.8$) to ($1.4, 0.9, 1.6, 0.2$) based on the timestep range [980,20], which are ranged by the original FreeU parameter values and inversely weighted values relative to 1.
The guidance weight for CFG is linearly mapped from a range of $100$ to $10$ over the iteration range [1,10,000].
The multi-view diffusion used in MVDream applies CFG with a guidance weight set to $50$, utilizing a negative prompt instead of unconditioned input~\cite{ban2024understandingimpactnegativeprompts}.
To prevent image over-exposure caused by large guidance weight, the Rescale CFG, which rescales after CFG, is applied following~\cite{shi2024mvdreammultiviewdiffusion3d}. 
Since the rescale method adjusts brightness rather than quality, it is excluded from dynamic scaling.
When FreeU and CFG dynamic scaling are applied, NeRF optimization takes about 2 hours on a single \textit{NVIDIA GeForce RTX 3090} GPU.

\subsection{Effect of dynamic scaling on multi-view SDS}
The 3D objects generated on \cref{fig:experiment} from each text prompt are displayed from three angles. 
The results highlight the application of training-free techniques, such as FreeU~\cite{si2023freeufreelunchdiffusion} and CFG~\cite{ho2022classifierfreediffusionguidance}, to the baseline method (MVDream~\cite{shi2024mvdreammultiviewdiffusion3d}), following the \textbf{effective dynamic scaling rules} to overcome the trade-offs inherent in simple scaling, which are detailed depiction and consistency for FreeU, and smoothness of surface and size for CFG.
When only FreeU dynamic scaling is harnessed (second row), it provides more detailed descriptions compared to the baseline (first row), successfully preventing artifacts (e.g., in the axe example) and geometric defects (e.g., in the corgi example). 
When only CFG dynamic scaling is harnessed (third row), it produces smoother surfaces than the baseline while maintaining the object size, as clearly demonstrated in the corgi example. 
Based on the analysis that each training-free techniques are applied at different stages in the SDS process, applying both scheduled FreeU and CFG allows us to harness the full potential of each technique, effectively improving the quality of text-to-3D outputs via score distillation.

\subsection{User study}
\begin{table}[ht]
    \centering
    \begin{tabular}{l c}
        \hline
        Method & User preference \% \\ \hline
        MVDream~\cite{shi2024mvdreammultiviewdiffusion3d} & 31.0 \\
        MVDream + Ours & \textbf{69.0} \\ \hline
    \end{tabular}
    \caption{\textbf{User study.} 
    A total of 200 responses comparing the overall quality with and without harnessing training-free techniques were analyzed. 
    Twice as many users preferred the results when our dynamic scaling scheme was used, over the baseline results.
    }
    \label{tbl:userstudy}
\end{table}
Since text-to-3D lacks ground-truth results according to text prompts, it is difficult to quantitatively evaluate the effectiveness of training-free techniques. 
Instead, we conducted a user study that compares results with and without the harnessing of such techniques.
The user study involved 20 participants evaluating 10 different prompts. 
As shown in \cref{tbl:userstudy}, more than twice as many participants judged that the results produced with the training-free techniques achieved higher quality outcomes.

\begin{figure*}
    \centering
    \includegraphics[width=\linewidth]{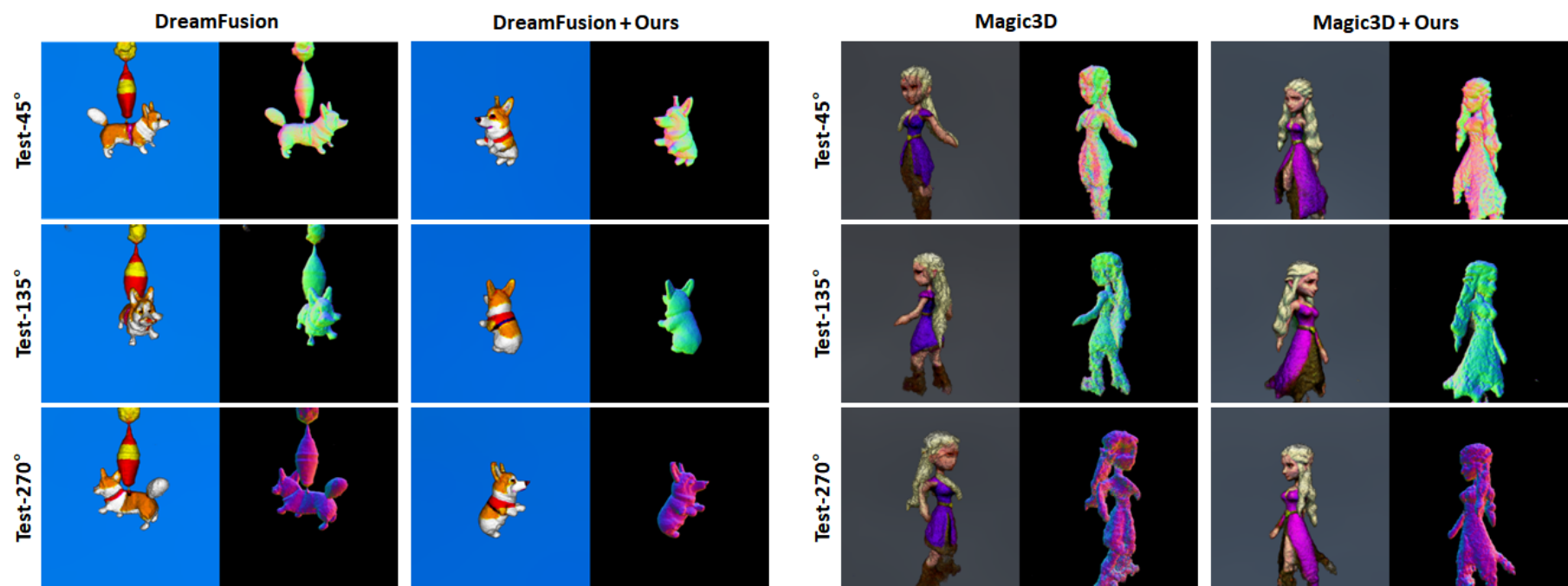}
    \caption{Effects of harnessing training-free techniques on \textbf{DreamFusion} and \textbf{Magic3D}}
    \label{fig:dreamfusion&magic3d}
    \vspace{-3.0mm}
\end{figure*}

\subsection{Generalization to other SDS-based methods}
To demonstrate the effectiveness of harnessing training-free techniques in SDS, experiments were conducted with other models that leverage score distillation, which are DreamFusion~\cite{poole2022dreamfusiontextto3dusing2d}, Magic3D~\cite{lin2023magic3dhighresolutiontextto3dcontent}.
For the same prompt, \textit{"Corgi riding a rocket"}, the left side of each section in \cref{fig:dreamfusion&magic3d} shows the results produced by the base models, and the right side shows the results after applying the dynamic training-free scaling techniques to each base model. 

Since both SDS-based models utilize diffusion models as a prior, the FreeU dynamic scaling technique can be applied. 
As illustrated in \cref{fig:dreamfusion&magic3d}, the harnessing of training-free techniques leads to more detailed outputs, and particularly the corgi's body is represented without any geometric defects.
Furthermore, regarding the effects of CFG dynamic scaling, the object size demonstrates uniform consistency regardless of dynamic scaling or not, and the 3D mesh in \cref{fig:dreamfusion&magic3d} shows that our dynamic scaling technique has the effect of making the surface of the object smooth.

Therefore, harnessing the power of training-free techniques such as FreeU and CFG is broadly effective in diffusion-based score distillation, and similar improvements can be consistently achieved.

\subsection{Effects of dynamic scaling rules on Image-to-3D}
\begin{figure}
    \centering
    \includegraphics[width=\linewidth]{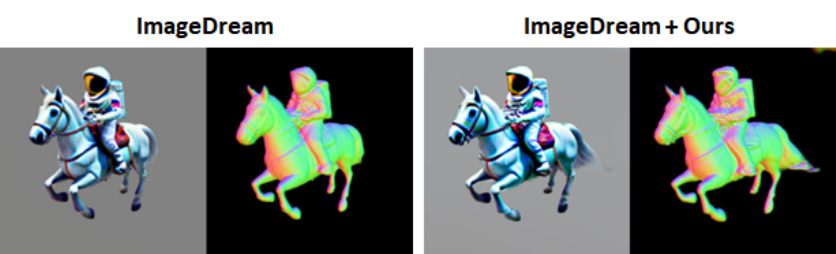}
    \caption{Results of applying our scheme on ImageDream~\cite{wang2023imagedreamimagepromptmultiviewdiffusion}}
    \label{fig:imagedream}
\end{figure}
We conduct additional experiments on score distillation for image-to-3D generation. 
Building on MVDream~\cite{shi2024mvdreammultiviewdiffusion3d}, ImageDream~\cite{wang2023imagedreamimagepromptmultiviewdiffusion} incorporates an image prompt to facilitate image-to-3D generation. 
As shown in \cref{fig:imagedream}, there is little difference between the results whether or not our dynamic scaling scheme is applied.
We conjecture that an image input serves as a very explicit and strong prior for the multi-view diffusion model, guiding it to produce outputs that closely resemble the input image prompt, mitigating the effect of FreeU or CFG.
This emphasizes the importance of using a good image prompt over other training-free techniques for text-to-3D generation.

\subsection{Failure cases}
\begin{figure}
    \centering
    \includegraphics[width=\linewidth]{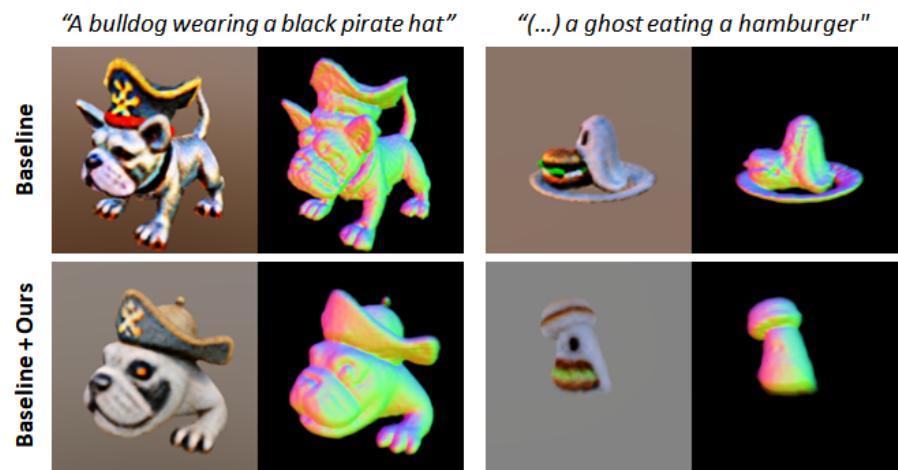}
    \caption{
    \textbf{Failure cases.}
    Our method sometimes generates incorrect shapes due to imbalances in the focus of the prompt (e.g., the prompt on the left prioritizes hat creation over the bulldog's hind legs) or overly broad descriptions (e.g., the prompt on the right requires more specific details about the ghost and hamburger).
    }
    \label{fig:failure}
\end{figure}
Dynamic scaling training-free techniques sometimes fail to improve quality. 
As illustrated in \cref{fig:failure}, it often occurs when the text prompt only focuses on certain aspects or provides insufficient detail.
Our dynamic scaling method amplifies the CFG scaling factor during the early stages of the NeRF optimization, which governs the generation of the shape. 
This approach enables a rapid convergence toward the intended direction of the reflecting the text prompt.
We hypothesize that this process leads to insufficient guidance for aspects not emphasized in the prompt.
Furthermore, when the prompt provides vague or incomplete information, we infer that the enhanced guidance may produce undesired shapes, such as a hamburger on the head of a ghost.
\section{Conclusion}
\label{sec:conclusion}

This paper demonstrates that dynamic scaling of the training-free techniques effectively enhances the quality of text-to-3D generation via score distillation sampling. 
Experiments were conducted with training-free scaling techniques such as classifier-free guidance (CFG) and FreeU, both of which efficiently improve quality in text-to-2D generation. 
CFG has a trade-off between surface smoothness and the size of 3D content, and FreeU also has a trade-off between detailed depiction and risk of geometric defects or artifacts in score distillation.
In order to strike a favorable balance between the trade-off relationships, and to maximize the positive effects on the generation output, we propose a strategic dynamic scaling scheme of such training-free techniques, which is applied at the appropriate stages within the score distillation process.
FreeU handles the diffusion UNet features as a prior, so it is scheduled according to the sampled timestep in an inversely proportional manner. 
In contrast, CFG manipulates the diffusion score, so it is scheduled according to the step of iterative optimization, also in an inversely proportional manner.
We demonstrate the effectiveness of our schemes through qualitative comparisons, a comprehensive user study, and CLIP scores of qualitative results.
We hope that this work motivates further research into effective training-free techniques for text-to-3D generation and Score Distillation Sampling.

\smallbreak
\noindent \textbf{Acknowledgements.}
This work was supported by the NRF grant (RS-2021-NR059830 (50\%)) and the IITP grants (RS-2021-II212068: AI Innovation Hub (45\%), RS-2019-II191906: Artificial Intelligence Graduate School Program at POSTECH (5\%)) funded by the Korea government (MSIT).
% \section{Acknowledgements}
% This work was supported by IITP grants (RS-2022-II220290:Visual Intelligence for Space-Time Understanding and Gen-eration (30\%), RS-2021-II212068: AI Innovation Hub(60\%), RS-2019-II191906: AI Graduate School Program at POSTECH (5\%), RS-2021-II211343: AI Graduate School Program at SNU: 2021-0-01343 (5\%)) funded by the Korea government.
\clearpage

{
    \small
    \bibliographystyle{ieeenat_fullname}
    \bibliography{main}
}

\clearpage
\setcounter{section}{0}
\setcounter{table}{0}
\setcounter{figure}{0}
\renewcommand{\thesection}{\Alph{section}}
\renewcommand\thefigure{A\arabic{figure}}
\renewcommand{\thetable}{A\arabic{table}}

\title{
Harnessing the Power of Training-Free Techniques in Text-to-2D Generation \\ for Text-to-3D Generation via Score Distillation Sampling
\\
{\it ----- Supplementary Material -----}
}
\maketitle

In this supplementary material, we offer further insights and qualitative results of harnessing the dynamic scaling that were omitted from the main paper due to space limitations.
~\cref{app:appendixA} provides additional implementation details and supplementary mathematical formulations of the theories used.
In ~\cref{app:appendixB}, the increased risk of inconsistency when applying the FreeU scaling technique to multi-view diffusion is clearly illustrated through figures, offering an intuitive understanding.
~\cref{app:appendixC} presents results from additional SDS-based text-to-3D models utilizing our dynamic scaling technique, which were not shown in the main paper. 
~\cref{app:appendixD} supports the necessity for dynamic scaling by displaying rendered outputs during the optimization process. 
Finally, ~\cref{app:appendixE} presents additional qualitative results to demonstrate the general applicability of the dynamic scaling method. 
For both the qualitative results presented in the main paper and those in ~\cref{app:appendixE}, the complete 360-degree views of the 3D content can also be observed through corresponding videos.

\section{Additional implementation details}\label{app:appendixA}
We use a representative open-source MVDream~\cite{shi2024mvdreammultiviewdiffusion3d} that utilizes a dataset composed of 70\% from the publicly available Objaverse dataset~\cite{deitke2022objaverseuniverseannotated3d} and 30\% from a subset of the LAION dataset~\cite{schuhmann2022laion5bopenlargescaledataset}, sampled in batches for training. 
For camera views, four random views that are mutually orthogonal are selected. 
The camera embedding is added as a residual to the time embedding and incorporated into the text embedding via cross-attention.

The model is optimized over 10,000 steps using the AdamW optimizer~\cite{kingma2017adammethodstochasticoptimization} with a learning rate of 0.01. 
In MVDream, timestep annealing is performed by randomly selecting a value within a range defined between the maximum and minimum timesteps. 
This range expands from [0.98, 0.98] to [0.02, 0.5] over the first 8,000 iterations, and after the 8,000th iteration, the range does not change.

\begin{equation}\label{eq:negativeprompt}
    x_{cfg} = x_{neg} + w(x_{pos} - x_{neg})
\end{equation}
The exact expression for classifier-free guidance~\cite{ho2022classifierfreediffusionguidance} with negative prompt~\cite{ban2024understandingimpactnegativeprompts} is shown in the above~\cref{eq:negativeprompt}, where $w$ is the guidance weight, $x_{pos}$ and $x_{neg}$ are the outputs using positive and negative prompts respectively. 
\begin{align}\label{eq:rescaleCFG}
    \sigma_{pos} = std(x_{pos}), \sigma_{cfg} = std(x_{cfg}) \nonumber \\
    x_{rescaled} = x_{cfg} \cdot \frac{\sigma_{pos}}{\sigma_{cfg}} \nonumber\\
    x_{final} = \phi \cdot x_{rescaled} + (1-\phi ) \cdot x_{cfg}
\end{align}

To prevent image over-exposure caused by large guidance weight, the Rescale CFG, which rescales after CFG, is applied following~\cite{shi2024mvdreammultiviewdiffusion3d}, as expressed in~\cref{eq:rescaleCFG}

\section{Additional visualization: Effect of FreeU on multi-view diffusion}\label{app:appendixB}
\begin{figure}
    \centering
    \includegraphics[width=\linewidth]{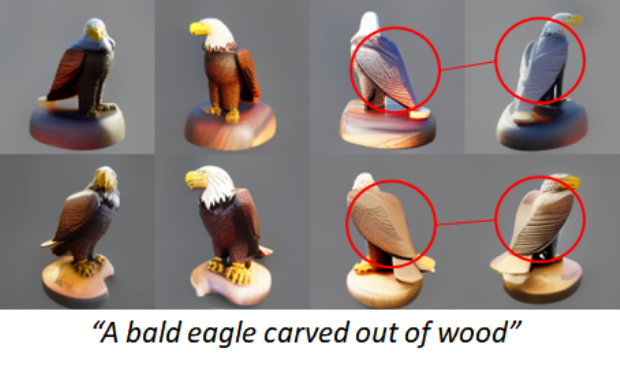}
    \includegraphics[width=\linewidth]{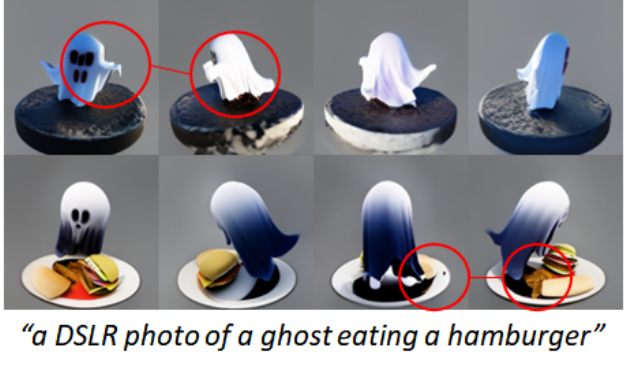}
    \includegraphics[width=\linewidth]{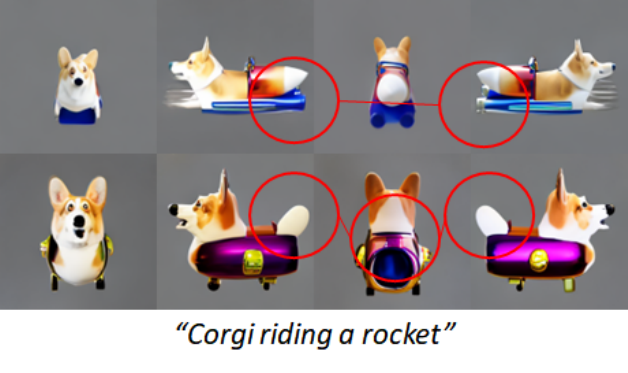}
    \includegraphics[width=\linewidth]{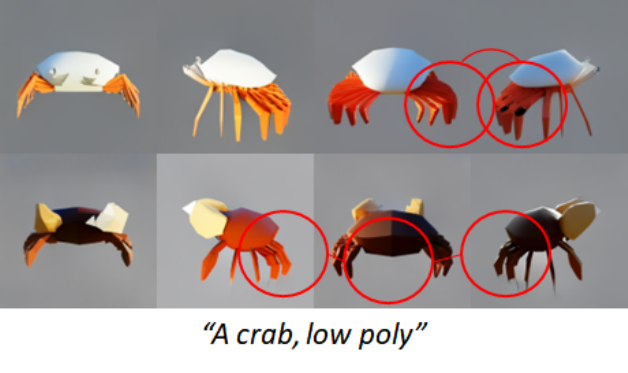}
    \caption{
    \textbf{Inconsistencies of multi-view diffusion with FreeU scaling}
    The results show that the multi-view diffusion, scaled by FreeU, generates two outputs for each of the four views.
    Although the same 3D object should maintain consistency across different views, the red circles highlight areas where inconsistencies have occurred. 
    FreeU scaling frequently and distinctly induces such inconsistencies. 
    The red circles indicate the inconsistencies that only occurred following the application of FreeU.
    }
    \label{fig:additionalmultiview}
\end{figure}
In multi-view diffusion, the FreeU scaling technique~\cite{si2023freeufreelunchdiffusion} presents a trade-off between capturing fine details and increasing the risk of inconsistencies. 
When the backbone feature scaling factor $b$ of FreeU is set to a larger value, object details are more accurately depicted, however, the inconsistencies also become more pronounced, as shown in~\cref{fig:additionalmultiview}. 
The figure displays outputs generated from the prompts, listed from top to bottom: ‘A bald eagle carved out of wood,’ ‘a DSLR photo of a ghost eating a hamburger,’ ‘Corgi riding a rocket,’ and ‘A crab, low poly.’

In the first set of results related to the \textit{eagle}, inconsistencies in the eagle's wing patterns are noticeable. 
In the second, concerning the \textit{ghost}, the direction of the ghost’s arms and the position of some snacks are changed.
The corgi prompt results in the appearance of additional accessories or the disappearance of its tail. 
Lastly, in the results linked to the crab, inconsistencies are observed in the crab’s legs.

To evaluate the effect of FreeU on multi-view diffusion, the parameters $b1$, $s1$, $b2$, and $s2$ were set to $1.4$, $0.9$, $1.6$, and $0.2$, respectively.

\section{Effect of our dynamic scaling on other SDS-based text-to-3D}\label{app:appendixC}
\begin{figure}
    \centering
    \includegraphics[width=\linewidth]{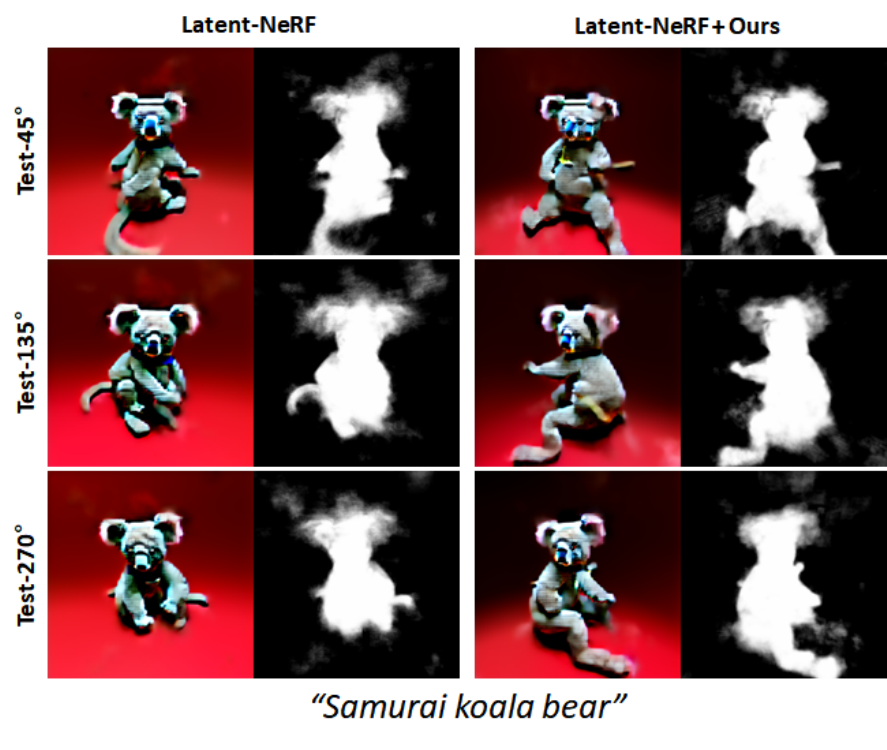}
    \includegraphics[width=\linewidth]{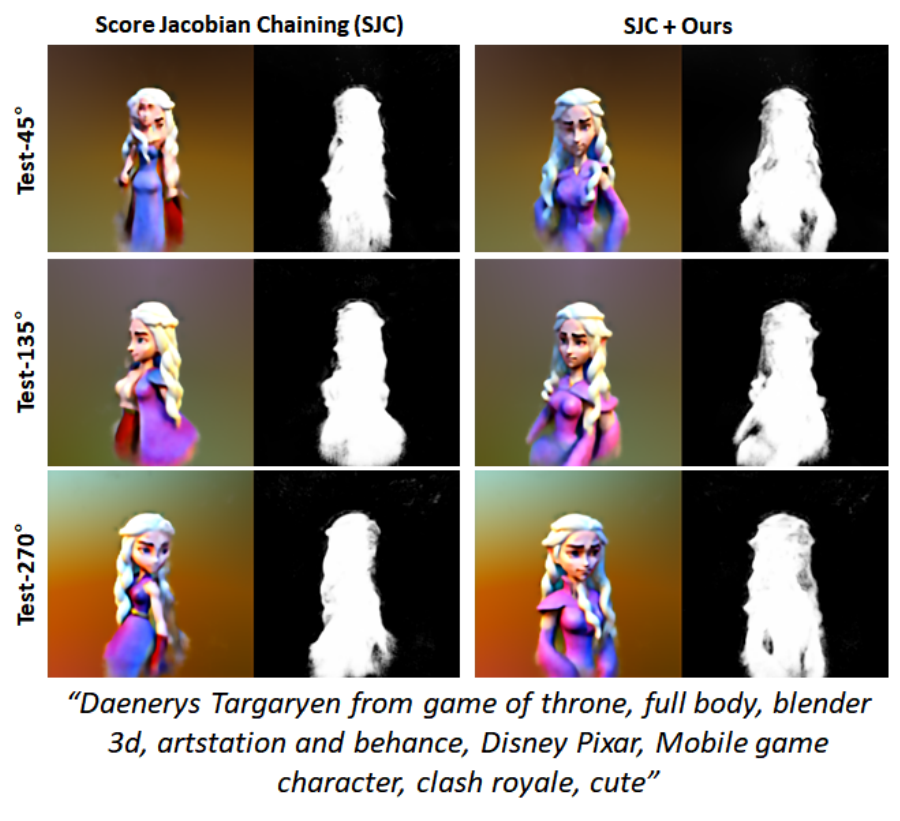}
    \caption{
    Effects of harnessing training-free techniques on \textbf{Latent-NeRF} and \textbf{Score Jacobian Chaining.}
    While the output does not include a mesh representation, making it difficult to assess the smoothness of the object's surface or the presence of geometric flaws, the effect of generating detailed and realistic 3D objects is still confirmed.
    }
    \label{fig:latentnerfandjacobianchaining}
\end{figure}
In the experiments harnessing dynamic scaling to SDS-based models, the outputs from Latent-NeRF~\cite{metzer2022latentnerfshapeguidedgeneration3d} and Score Jacobian Chaining (SJC)~\cite{wang2022scorejacobianchaininglifting} in~\cref{fig:latentnerfandjacobianchaining} do not provide mesh representations, unlike DreamFusion~\cite{poole2022dreamfusiontextto3dusing2d} and Magic3D~\cite{lin2023magic3dhighresolutiontextto3dcontent}. 
The absence of mesh output makes it difficult to evaluate the effects of CFG scaling in a dynamic manner on object surface smoothness and to assess the geometric flaws from the dynamic FreeU scaling technique, which are otherwise easily observable in mesh form. 
However, as demonstrated in~\cref{fig:latentnerfandjacobianchaining}, our dynamic scaling technique effectively enhances detail representation while maintaining object size.

The figure showcases the results generated from the prompts `Samurai koala bear' and `Daenerys Targaryen from Game of Thrones, full body, Blender 3D, Artstation and Behance, Disney Pixar, Mobile game character, Clash Royale, cute.' In the output of Latent-NeRF for the koala prompt, the model produced a realistic and coherent koala with added details, such as the samurai sword. In the example for SJC, there is a noticeable increase in the facial detail of the character.

\section{CLIP score}
The CLIP score is a metric that measures text-image similarity using CLIP (Contrastive Language-Image Pretraining)~\cite{radford2021learning}, a model trained on approximately 400 million text-image pairs through contrastive pretraining. 
However, since our technique specifically targets text-to-3D generation via score distillation sampling, the CLIP score is not well-suited for rigorously evaluating the effectiveness of our techniques. 
In particular, the CLIP score can significantly depend on the viewing angle of the same generated 3D object, making it an unreliable indicator of the quality of the generation.
However, the CLIP scores computed for the rendered images presented in the figure are consistent with and supportive of our findings.

In the context of text-to-3D generation via score distillation sampling, classifier-free guidance (CFG), which aims to better align the output with the text condition, results in a trade-off between the object’s size and the smoothness of its surface.
The CLIP scores for the corresponding examples shown in ~\cref{fig:CFG_SDS} are as follows:
\setlength{\textfloatsep}{5pt}
\begin{table}[ht]
\centering
\resizebox{\columnwidth}{!}{
\begin{tabular}{lcccc}
\toprule
\textbf{Viewing angle (°)} & \textbf{Baseline} & \textbf{Small CFG} & \textbf{Large CFG} & \textbf{Dynamic CFG} \\
\midrule
45   & 0.3329 & 0.3074 & 0.3180 & 0.3356 \\
135  & 0.2450 & 0.2677 & 0.2505 & 0.2418 \\
270  & 0.2908 & 0.2692 & 0.2542 & 0.3164 \\
\midrule
\textbf{maximum score} & 0.3329 & 0.3074 & 0.3180 & 0.3356 \\
\bottomrule
\end{tabular}
}
\caption{CLIP scores under different CFG settings and viewing angles which are corresponding to ~\cref{fig:CFG_SDS}}
\label{tbl:CLIPforCFG}
\end{table}

\noindent
As presented in ~\cref{tbl:CLIPforCFG}, when the generated object is relatively small or exhibits a rough surface, the CLIP score tends to decrease compared to the baseline result.
However, with the application of our dynamic scaling technique, this issue is mitigated, as evidenced by improved CLIP scores in such cases.

\begin{table}[ht]
\centering
\resizebox{\columnwidth}{!}{
\begin{tabular}{lcccc}
\toprule
\textbf{Viewing angle (°)} & \textbf{Baseline} & \textbf{FreeU scaling b,s} & \textbf{FreeU scaling b} & \textbf{Dynamic FreeU} \\
\midrule
45   & 0.2534 & 0.2723 & 0.2825 & 0.3019 \\
135  & 0.2637 & 0.2752 & 0.2967 & 0.2640 \\
270  & 0.2529 & 0.2873 & 0.2929 & 0.2577 \\
\midrule
\textbf{maximum score} & 0.2637 & 0.2873 & 0.2967 & 0.3019 \\
\bottomrule
\end{tabular}
}
\caption{CLIP scores under different FreeU settings and viewing angles which are corresponding to ~\cref{fig:FreeU_SDS}}
\label{tbl:CLIPforFreeU}
\end{table}

\begin{table}[ht]
\centering
\resizebox{\columnwidth}{!}{
\begin{tabular}{lcccc}
\toprule
\textbf{Viewing angle (°)} & \textbf{Baseline} & \textbf{Dynamic FreeU} & \textbf{Dynamic CFG} & \makecell[c]{\textbf{Dynamic FreeU+CFG} \\ \textbf{(ours)}} \\
\midrule
45   & 0.2167 & 0.2219 & 0.2325 & 0.2351 \\
135  & 0.2196 & 0.2410 & 0.2073 & 0.2647 \\
270  & 0.2270 & 0.2388 & 0.2366 & 0.2550 \\
\midrule
\textbf{maximum score} & 0.2270 & 0.2410 & 0.2366 & 0.2647 \\
\bottomrule
\end{tabular}
}
\caption{CLIP scores for the top of ~\cref{fig:experiment}. }
\label{tbl:CLIPforExp}
\end{table}

Shown in ~\cref{tbl:CLIPforFreeU}, the enhancement of texture details resulting from the amplification of backbone features via FreeU is also reflected in the CLIP scores.
Furthermore, the minimal difference in CLIP scores between the settings with and without skip feature scaling indicates that this component has a limited influence on multi-view diffusion.

As demonstrated in ~\cref{tbl:CLIPforExp}, dynamic FreeU reduces the risk of geometric errors while enhancing texture details, resulting in higher CLIP scores compared to the baseline.
Dynamic CFG controls the trade-off between object size and surface smoothness, leading to improved scores over the baseline with static CFG scaling.
As these two techniques target different components of the generation pipeline and are functionally independent, our combined approach achieves the highest CLIP scores.

\section{Progressive visualization and analysis of 3D generation}\label{app:appendixD}
\begin{figure*}
    \centering
    \includegraphics[width=\linewidth]{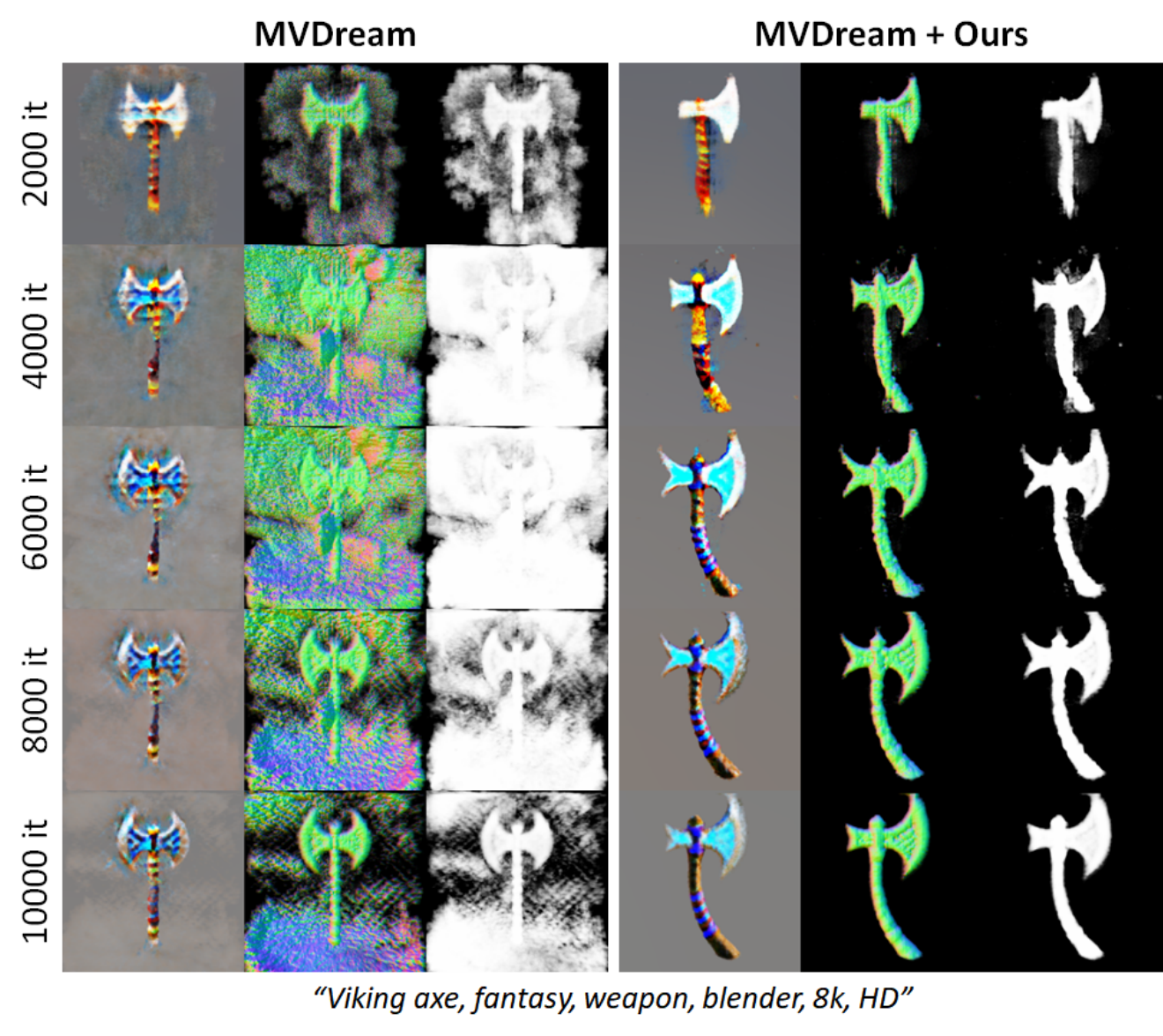}
    \caption{
    \textbf{Visualization of rendered outputs during NeRF optimization.}
    Our strategic dynamic scaling method adjusts the CFG by reducing the guidance weight and modifies the FreeU by strengthening the backbone feature as optimization iterations increase and timesteps decrease, respectively. 
    The effect of training-free techniques such as CFG and FreeU is twofold: in the early stages, they help mitigate errors like geometric defects and artifacts, while in the later stages, they contribute to smoother surfaces and more refined detail representation. 
    This visualization illustrates the rendering process of 3D content generated from the prompt `Viking axe, fantasy, weapon, blender, 8k, HD'.
    When comparing the intermediate stages of optimization, the superiority of our dynamic scaling technique becomes clearly evident.
    }
    \label{fig:duringoptimization}
\end{figure*}
\begin{figure*}
    \centering
    \includegraphics[width=\linewidth]{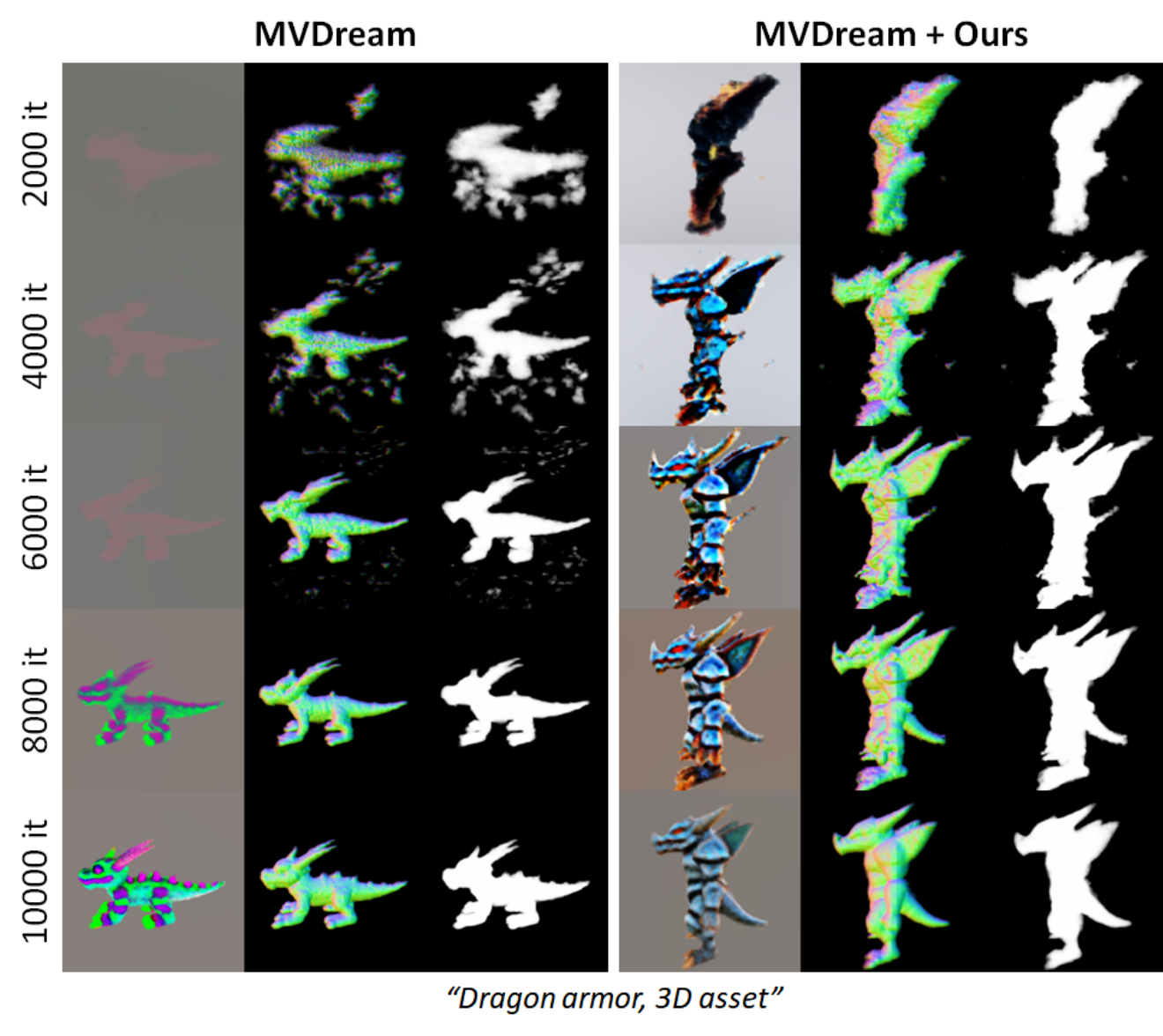}
    \caption{
    \textbf{Visualization of rendered outputs during NeRF optimization.}
    Our strategic dynamic scaling method adjusts the CFG by reducing the guidance weight and modifies the FreeU by strengthening the backbone feature as optimization iterations increase and timesteps decrease, respectively. 
    The effect of training-free techniques such as CFG and FreeU is twofold: in the early stages, they help mitigate errors like geometric defects and artifacts, while in the later stages, they contribute to smoother surfaces and more refined detail representation. 
    This visualization illustrates the rendering process of 3D content generated from the prompt `Dragon armor, 3D asset'.
    When comparing the intermediate stages of optimization, the superiority of our dynamic scaling technique becomes clearly evident.
    }
    \label{fig:duringoptimization}
\end{figure*}
\begin{figure*}
    \centering
    \includegraphics[width=\linewidth]{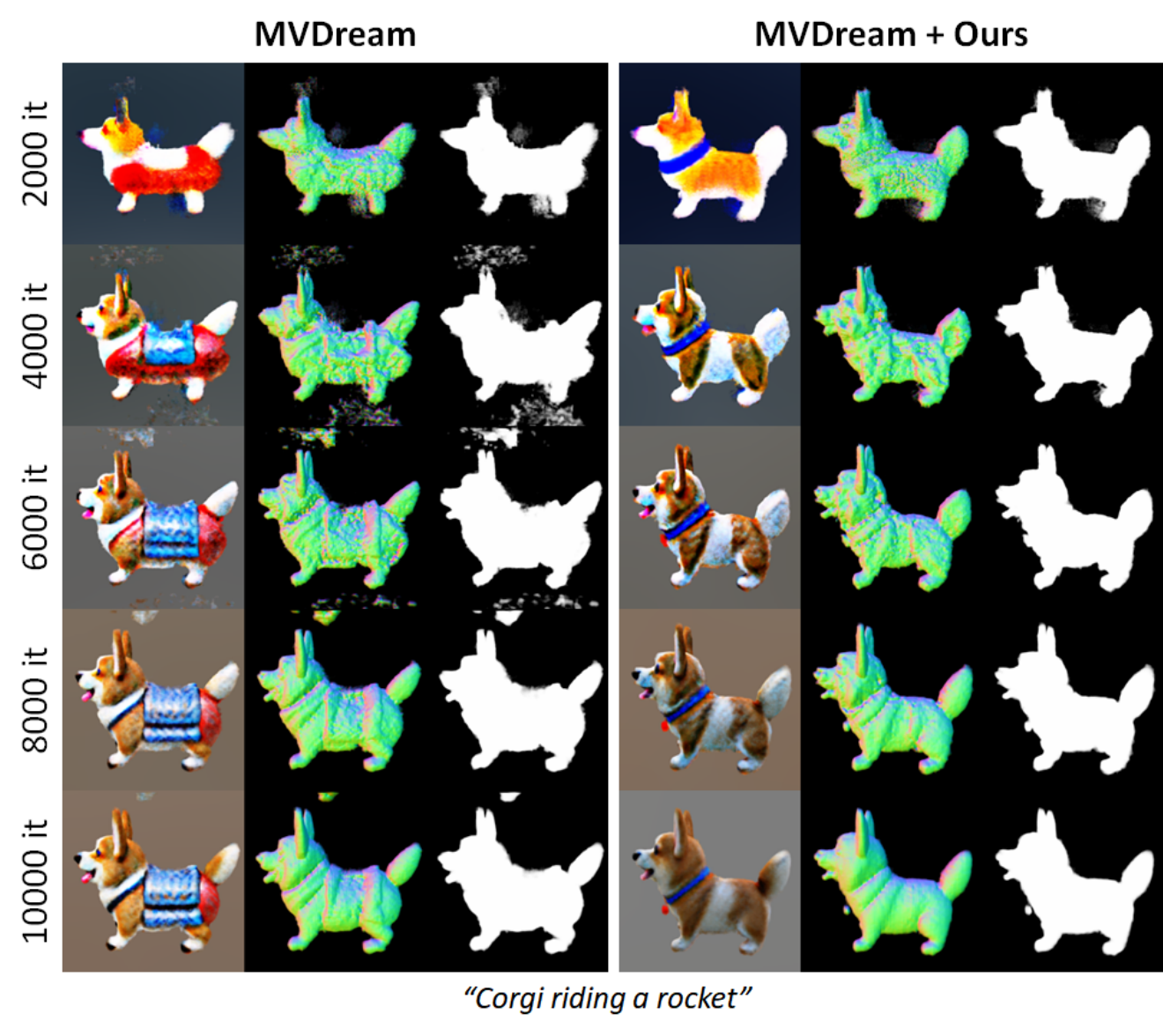}
    \caption{
    \textbf{Visualization of rendered outputs during NeRF optimization.}
    Our strategic dynamic scaling method adjusts the CFG by reducing the guidance weight and modifies the FreeU by strengthening the backbone feature as optimization iterations increase and timesteps decrease, respectively. 
    The effect of training-free techniques such as CFG and FreeU is twofold: in the early stages, they help mitigate errors like geometric defects and artifacts, while in the later stages, they contribute to smoother surfaces and more refined detail representation. 
    This visualization illustrates the rendering process of 3D content generated from the prompt `Corgi riding a rocket'.
    When comparing the intermediate stages of optimization, the superiority of our dynamic scaling technique becomes clearly evident.
    }
    \label{fig:duringoptimization}
\end{figure*}
The dynamic scaling method proposed in this paper reduces the CFG guidance weight as optimization iterations increase, while simultaneously increasing the FreeU backbone feature scaling factor as timesteps decrease. 
Through timestep, annealing~\cite{shi2024mvdreammultiviewdiffusion3d}, optimization iterations and timesteps are generally inversely proportional. 
In the early stages of optimization, the dynamic scaling technique reduces the risk of geometric defects or artifacts via FreeU, while CFG helps maintain object size. 
In the later stages, FreeU enhances the detail representation, and CFG ensures smoother surfaces.

In~\cref{fig:duringoptimization}, the early-stage results show that, compared to the baseline, our method significantly suppresses artifacts. 
The generated corgi retains a similar size but with less deformation, particularly in the back, where there is less noticeable sinking.
In the later stages of~\cref{fig:duringoptimization}, the output displays a significantly smoother surface than the results of MVDream, with realistic shading and detailing, including the intricate depiction of the bell attached to the corgi’s collar.

\section{Additional qualitative results}\label{app:appendixE}
\begin{figure*}
    \centering
    \includegraphics[width=0.825\linewidth]{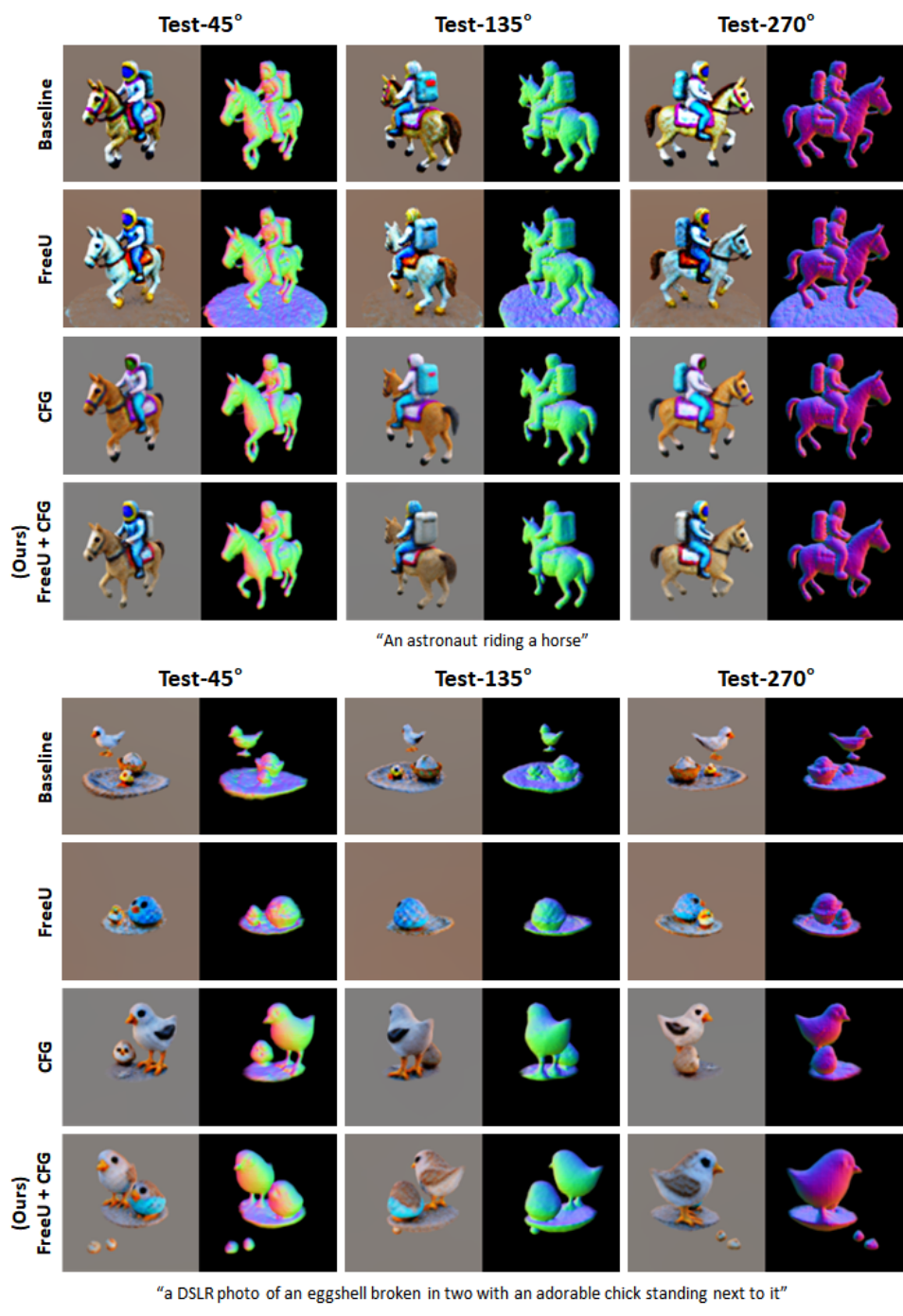}
    \caption{
    \textbf{Samples generated by score distillation with or without dynamic scaling training-free techniques.}
    }
    \label{fig:additional1}
\end{figure*}
\begin{figure*}
    \centering
    \includegraphics[width=0.825\linewidth]{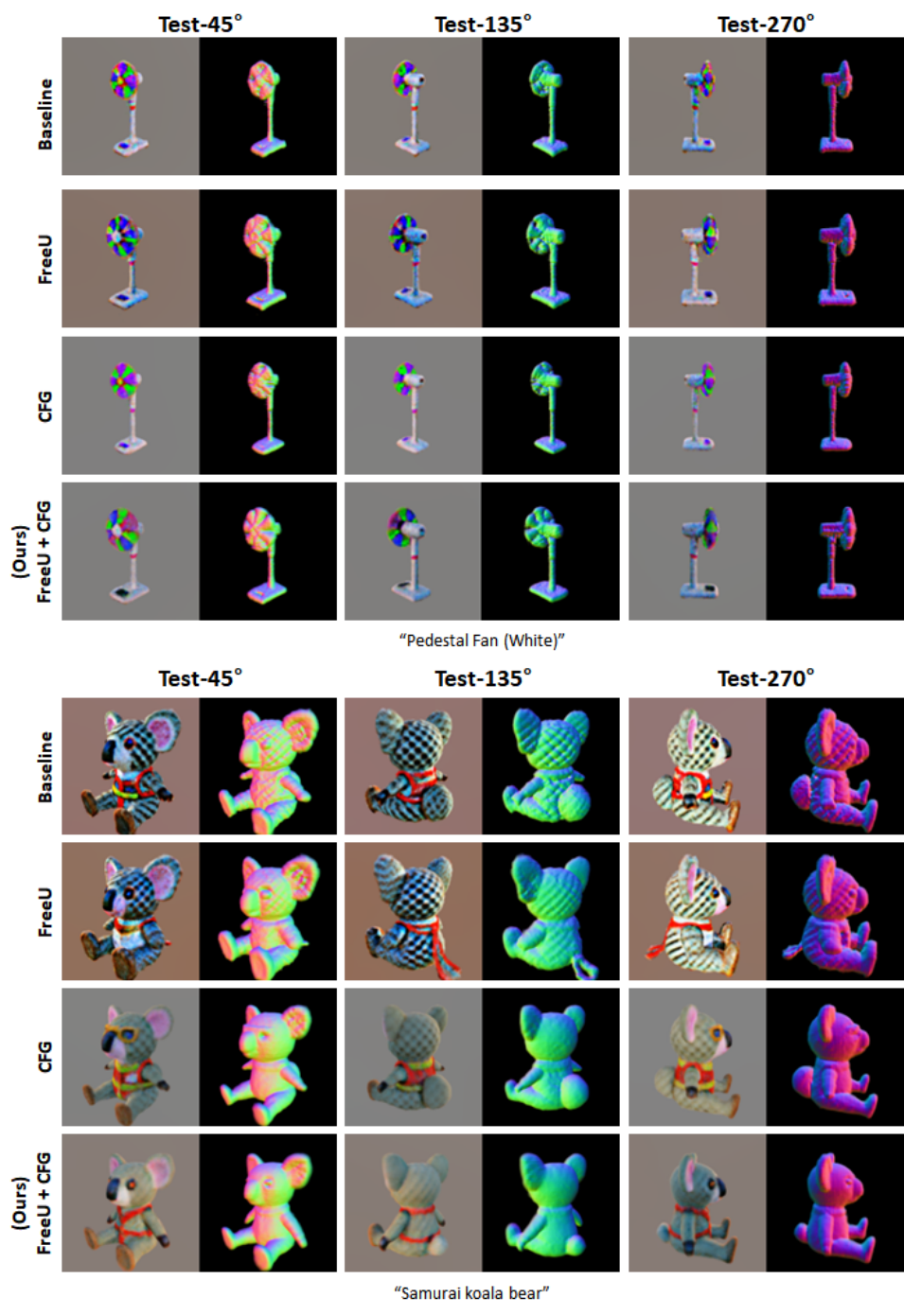}
    \caption{
    \textbf{Samples generated by score distillation with or without dynamic scaling training-free techniques.}
    }
    \label{fig:additional2}
\end{figure*}
\begin{figure*}
    \centering
    \includegraphics[width=0.825\linewidth]{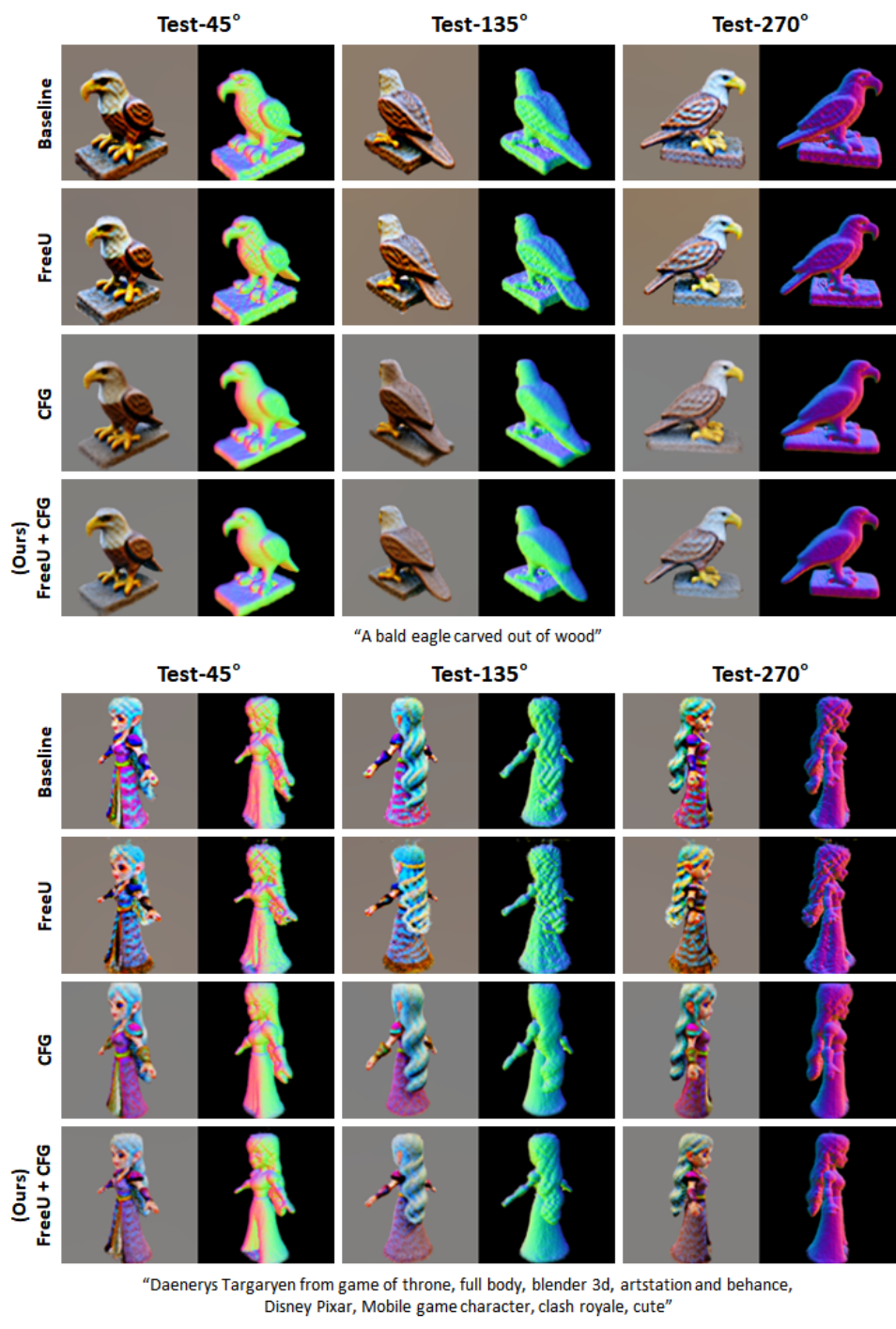}
    \caption{
    \textbf{Samples generated by score distillation with or without dynamic scaling training-free techniques.}
    }
    \label{fig:additional3}
\end{figure*}
For qualitative comparison, the representative open-source implementation of multi-view diffusion, MVDream~\cite{shi2024mvdreammultiviewdiffusion3d}, was used as the baseline. 
Dynamic FreeU scaling enhances detail representation and suppresses geometric errors, while dynamic CFG scaling contributes to smoother surface generation and maintains object size.
The results from additional prompts further demonstrate the quality improvements achieved through \textbf{dynamic scaling}, as shown in the figures. 
~\cref{fig:additional1} illustrates the outputs for the prompts `An astronaut riding a horse' and `a DSLR photo of an eggshell broken in two with an adorable chick standing next to it'. 
~\cref{fig:additional2} presents the results for `Pedestal Fan (White)' and `Samurai koala bear,' while ~\cref{fig:additional3} features `A bald eagle carved out of wood' and `Daenerys Targaryen from Game of Thrones, full body, Blender 3D, Artstation and Behance, Disney Pixar, Mobile game character, Clash Royale, cute.'

The additional prompts were selected based on evaluations from a user study, where many participants judged that the results produced using the dynamic scaling techniques demonstrated higher quality outcomes.
The enhanced 3D content generated by the dynamic scaling method can also be observed through 360-degree rotating videos.

\end{document}